\documentclass[letterpaper, 10 pt, conference]{ieeeconf}

\IEEEoverridecommandlockouts

\usepackage{url}
\usepackage{multicol}
\usepackage[bookmarks=true]{hyperref}
\usepackage{soul}
\usepackage{color}
\usepackage{flushend}
\usepackage{times}
\usepackage{multicol}
\usepackage{graphicx}
\usepackage{lipsum}
\usepackage{setspace}
\usepackage{epsfig}
\usepackage{mathptmx}
\usepackage{amsmath}
\usepackage{bm}
\usepackage{bbold}
\usepackage{amssymb}
\usepackage{tabularx}
\usepackage{mathtools}
\usepackage{color,soul}
\usepackage{paralist}
\usepackage{tikz}
\usepackage{float}
\usepackage{pdfpages}
\usepackage{makecell}
\usepackage[colorinlistoftodos,prependcaption,textsize=smallsize]{todonotes}
\usepackage{array}
\usepackage{xcolor}

\newcolumntype{L}[1]{>{\raggedright\let\newline\\\arraybackslash\hspace{0pt}}m{#1}}
\newcolumntype{C}[1]{>{\centering\let\newline\\\arraybackslash\hspace{0pt}}m{#1}}
\newcolumntype{R}[1]{>{\raggedleft\let\newline\\\arraybackslash\hspace{0pt}}m{#1}}

\newcommand{\inv}{^{-1}}
\newcommand{\tr}{^{\!\top}}

\newcommand{\argmin}{\operatornamewithlimits{argmin}}

\newcommand{\SE}{SE(3) }
\newcommand{\SO}{SO(3)}
\newcommand{\R} {{\rm I\!R}}

\newcommand{\bl}{{\bar l}}

\title{\LARGE \bf
	Dynamic SLAM: The Need For Speed }

\author{Mina Henein$^{1}$, Jun Zhang$^{1}$, Robert Mahony$^{1}$ and Viorela Ila$^{2}$
	\thanks{$^{1}$Mina Henein, Jun Zhang, and Robert Mahony are with the College of Engineering and Computer Science,
		Australian National University, 0200 Canberra, Australia.
		{\tt\small firstname.lastname@anu.edu.au}}%
	\thanks{$^{2}$Viorela Ila is with the School of Aerospace, Mechanical and Mechatronic Engineering,
		University of Sydney, 2006 Sydney, Australia
		{\tt\small viorela.ila@sydney.edu.au}}%
}

\begin{document}
		
	\maketitle
	\thispagestyle{empty}
	\pagestyle{empty}

\begin{abstract}
The static world assumption is standard in most simultaneous localisation and mapping (SLAM) algorithms.
Increased deployment of autonomous systems to unstructured dynamic environments is  driving a need to identify moving objects and estimate their velocity in real-time.
Most existing SLAM based approaches rely on a database of $3$D models of objects or impose significant motion constraints.
In this paper, we propose a new feature-based, model-free, object-aware dynamic SLAM algorithm that exploits semantic segmentation to allow estimation of motion of rigid objects in a scene without the need to estimate the object poses or have any prior knowledge of their $3$D models.
The algorithm generates a map of dynamic and static structure and has the ability to extract velocities of rigid moving objects in the scene.
Its performance is demonstrated on simulated, synthetic and real-world  datasets.
\end{abstract}
\section{Introduction}
\label{sec:Intro}
%

SLAM is an established research field in robotics.
While many accurate and efficient solutions to the problem exist, most of the existing techniques heavily rely on the static world assumption~\cite{Walcott12icra}.
This assumption limits the deployment of existing algorithms to a wide range of increasingly important real world scenarios involving dynamic and unstructured environments.
Advances in deep learning have provided algorithms that can reliably detect and segment classes of objects at almost real time~\cite{Detectron2018,He17iccv}.
To incorporate such information in a geometric SLAM formulation then either a 3D-model of the object must be available~\cite{Salas13cvpr,Galvez16ras} or the front end must explicitly provide pose information in addition to detection and segmentation \cite{MOT16,Byravan17icra,Wohlhart15cvpr}.
The requirement for accurate 3D-models severely limits the potential domains of application, while to the best of our knowledge, multiple object tracking and $3$D pose estimation remain a challenge to learning techniques.
There is a clear need for an algorithm that can exploit the powerful detection and segmentation capabilities of modern deep learning algorithms without relying on additional pose estimation or motion model priors.

\begin{figure}[t]
	\centering
	\includegraphics[width=0.9\linewidth,trim=2cm 0cm 0cm 0cm,clip]{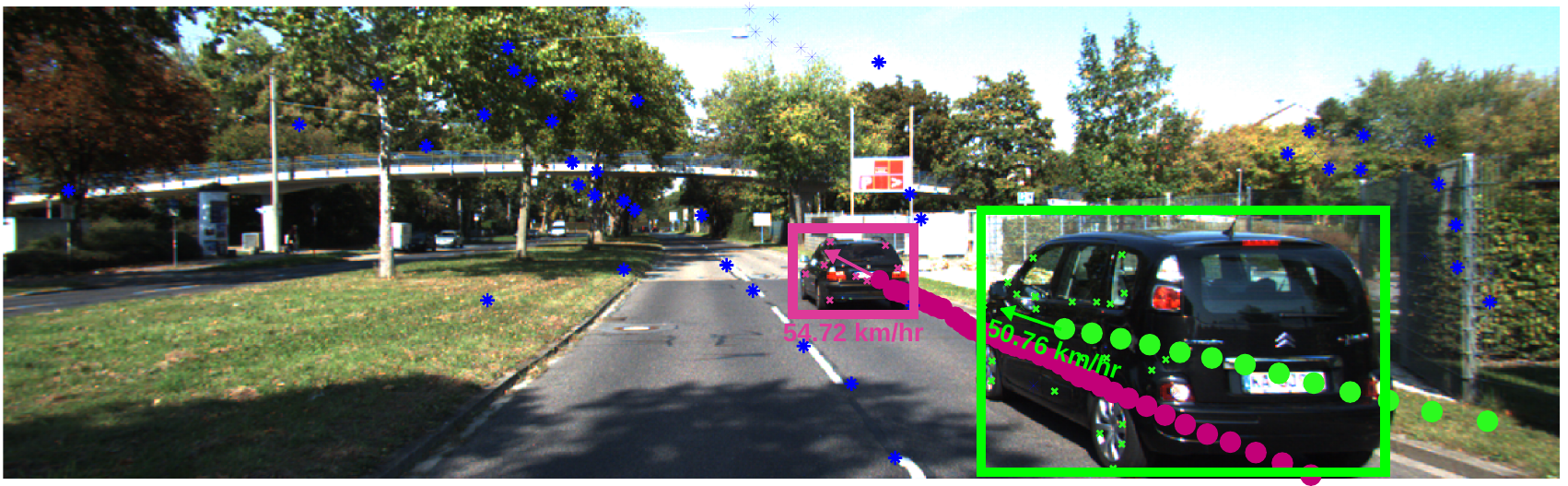}
	\caption{\scriptsize Results of our object-aware dynamic SLAM on KITTI Sequence0003. Centroids of each object are obtained by applying our motion estimates to the first ground-truth object centroid. Speed estimates are also extracted for each object.}
	\label{fig:kitti3-speed}
\end{figure}
%

In this paper, we propose a novel model-free, object-aware  point-based dynamic SLAM approach that leverages image-based semantic information to simultaneously localise the robot, map the static structure, estimate a full \SE pose change of moving objects and build a dynamic representation of the world.
We also fully exploit the rigid object motion to extract velocity information of objects in the scene (Fig.~\ref{fig:kitti3-speed}), an emerging task in autonomous driving which has not yet been thoroughly explored~\cite{Kampelmuhler18cvww}.
Such information is crucial to aid autonomous  driving algorithms for tasks such as collision avoidance~\cite{Aufrere03mechatronics} and adaptive cruise control~\cite{Jurgen06sae}.
The key innovation in the paper is a novel \emph{pose change representation} used to model the motion of a collection of points pertaining to a given rigid body and the integration of this model into a SLAM optimisation framework.
The resulting algorithm is agnostic to the underlying 3D-model of the object as long as the semantic detection and segmentation of the object can be tracked.
To the best of our knowledge, this is the first work able to estimate, along with the camera poses, the static and dynamic structure, the full \SE pose change of every rigid object in the scene, extract object velocities and be demonstrable on a real-world outdoor dataset.

\section{Related Work}
\label{sec:Litreview}
Establishing the spatial and temporal relationships between a robot, stationary and moving objects in a scene serves as a basis for scene understanding~\cite{Wang07ijrr} and the problems of simultaneous localisation, mapping and moving object tracking are mutually beneficial. 
In the SLAM community, information associated with stationary objects is considered positive, while information drawn from moving objects is seen as degrading the algorithm performance. 
SLAM systems either treat data from moving objects as outliers~\cite{Hahnel02iros,Hahnel03icra,Wolf05autonrobot,Zhao08icra,Bescos18arXiv} or they track them separately using multi-target tracking~\cite{Wang03icra,Miller07icra,Rogers10iros,Kundu11iccv}. 
%
Bibby and Reid's SLAMIDE~\cite{bibby07rss} estimates the state of 3D features (stationary or dynamic) with a generalised EM algorithm where they use reversible data association to include dynamic objects in a single framework SLAM.
Wang \emph{et al.}~\cite{Wang07ijrr} developed a theory for performing SLAM with Moving Objects Tracking (SLAMMOT).
In the latest version of their SLAM with detection and tracking of moving objects, the estimation problem is decomposed into two separate estimators (moving and stationary objects) to make it feasible to update both filters in real time.
Kundu \emph{et al.}~\cite{Kundu11iccv} tackle the SLAM problem with dynamic objects by solving the problems of Structure from Motion (SfM) and tracking of moving objects in parallel. 
Reddy \emph{et al.}~\cite{Reddy15iros} uses optical flow and depth to compute semantic motion segmentation. 
They isolate static objects from moving objects and reconstruct them independently, before using semantic constraints to improve the the 3D reconstruction.
Dewan \emph{et al.}\cite{Dewan16icra} presents a model-free approach for detecting and tracking dynamic objects in 3D using LiDAR scans.
Judd \emph{et al.} ~\cite{Judd18iros} estimates the full \SE motion of both the camera and rigid objects in the scene by applying a multi-motion visual odometry (MVO) multimodel fitting technique.
Although this approach does not require prior knowledge of the environment or object $3$D models, they parameterise the motion transforms non-incrementally (with respect to the first observed frame) which might introduce severe linearisation errors and  only show results on one lab-environment experiment, with no evaluation on any existing datasets.
A very recent work by Yang and Scherer~\cite{Yang19tro} presents a method for single image 3$D$ cuboid detection, and multi-view object SLAM for both static and dynamic environments. 
Their main interest, however, is the camera pose and object detection accuracy and they provide no evaluation of the object pose estimation.


\section{Accounting For Dynamic Objects In SLAM}
\label{sec:Problem}

\begin{figure}[t]
	\begin{minipage}[t]{0.2\textwidth}
		\begin{minipage}[b]{\textwidth}
			\centering
			\includegraphics[height=3cm,trim=50mm 180mm 70mm 40mm,clip]{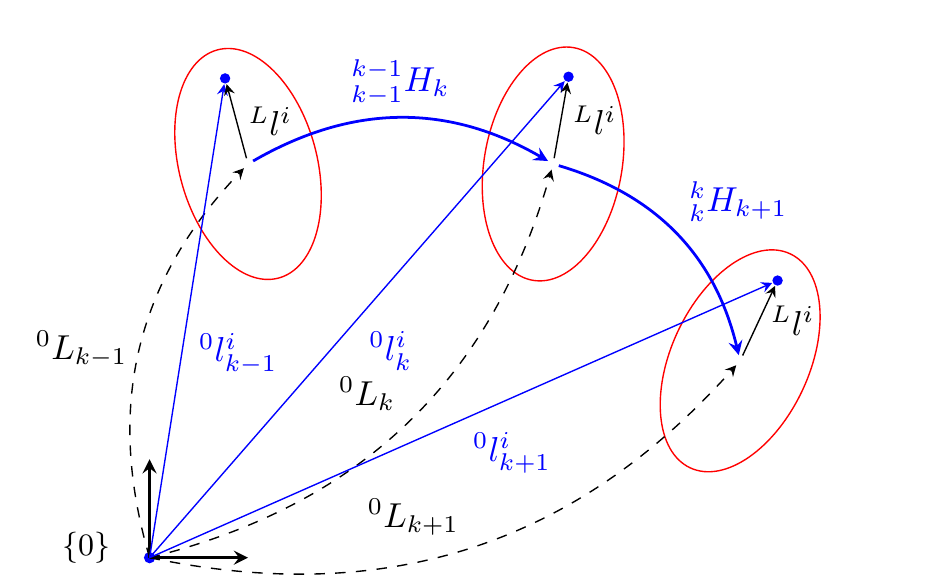}
			\\(a)
		\end{minipage}
	\end{minipage}\hspace{0.8cm}
	\begin{minipage}[t]{0.2\textwidth}
		\begin{minipage}[b]{\textwidth}
			\centering
			\includegraphics[height=3cm,trim=50mm 165mm 70mm 45mm,clip]{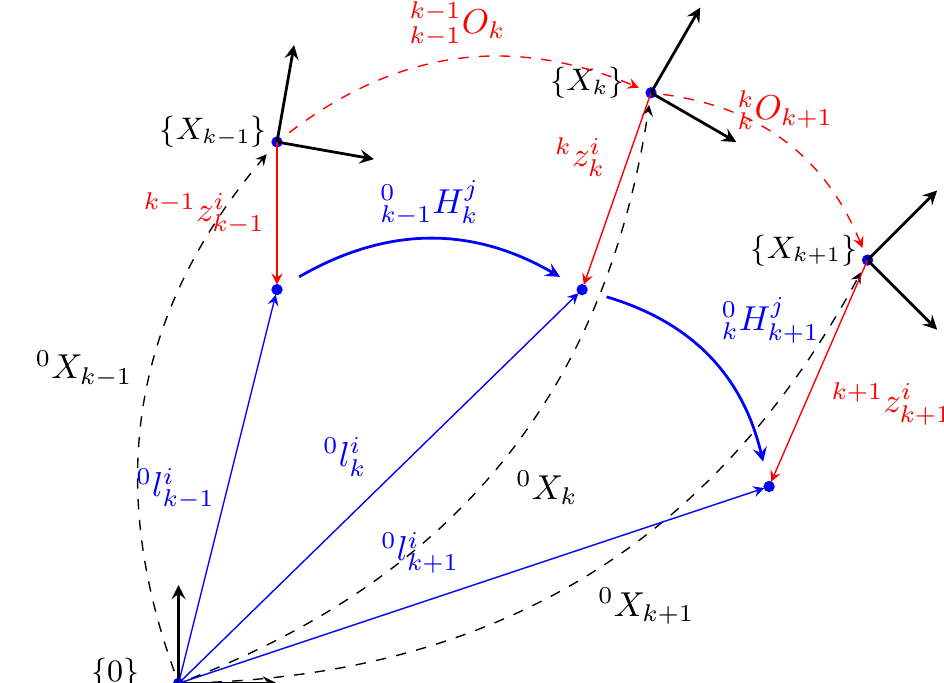}
			\\(b)
		\end{minipage}
	\end{minipage}
	\caption{\scriptsize (a): Coordinates of the rigid body in motion. The points $^Ll^i$ are represented relative to the rigid body pose $\{L\}$ at each step. (b): Robot poses, moving landmark positions (black) and the measurements (red) at three different time steps.} 
	\vspace{-.5cm}
\label{fig:Coordinates}
\end{figure}

The problem considered is one in which there are relatively large rigid objects moving within the sensing range of the robot that is undertaking the SLAM estimation.
The SLAM front-end is able to identify and associate points from the same potentially moving object at different time steps. These points share an underlying motion constraint that can be exploited to improve the quality of the SLAM estimation.
\subsection{Problem Formulation}
The SLAM with dynamic objects estimation problem is modelled using factor graphs~\cite{Dellaert06ijrr}, and the goal is to obtain the static and dynamic 3$D$ structure and the robot poses that maximally satisfy a set of measurements and pose change constraints. Assuming Gaussian noise, this problem becomes a non-linear least squares (NLS) optimisation over a set of variables~\cite{Polok13icra}: the robot poses \mbox{$\textbf{x} = \{x_0 ... x_{n_x}\}$}, with $x_k\in \SE$ 
where $k \in 0 ... n_x$ and $n_x$ is the number of steps, and the 3D point features in the environment seen at different time steps:  \mbox{${\bf l} = \{l^1_0 \dots \:l^{n_l}_{n_x}\}$} where \mbox{$l^i_k \in \R^3$ } and $i \in 1 ... n_l$ is the unique landmark index and $n_l$ is the total number of detected landmarks.
The set of landmarks, ${\bf l} = {\bf l_s} \cup {\bf l_d}$, contains a set of static landmarks ${\bf l_s}$ and a set of moving object landmarks ${\bf l_d}$. The same point on a moving object is represented using a different variable at each time step, i.e. $l^i_{k-1}$ and $l^i_k$ are the same physical $i^{th}$ point seen at times $k-1$ and $k$, respectively.

\subsection{Motion Model Of A Point On A Rigid Body}
Let $\{0\}$ denote the \emph{reference coordinate frame}, and $\{L\}$ the coordinate frame associated to a moving rigid body.
We write the pose $^0 L_k\in \SE$ of the rigid-body with respect to the reference frame $\{0\}$.
For a feature observed on an object, let ${^L} l^i \in \R^3$ denote the coordinates of this point in the object frame.
We write ${^0}l^i_k$ for the coordinates of the same point expressed in the reference frame $\{0\}$ at time $k$.
Note that for rigid bodies in motion, ${^L} l^i$ is constant for all the object instances, while both $^0 L_k$ and ${^0}l^i_k$ are time varying.
The point coordinates are related by the expression:
\begin{eqnarray}
^L\bl^i =\:  ^0 L_k\inv \: ^0\bl_k^i \label{eq:TransPoint}
\label{eq:RelPoint}
\end{eqnarray}
where the bar indicates homogeneous coordinates.

The relative motion of the object $L$ from $k-1$ to $k$ is represented by a rigid-body transformation \mbox{$\prescript{L_{k-1}}{k-1}H^{}_{k}\in \SE$} called the \emph{body-fixed frame  pose change}.
The indices indicate that the transformation maps a base pose $^0L_{k-1}$ (lower left index) to a target pose $^0L_{k}$ (lower right index), expressed in coordinates of the frame $^0L_{k-1}$ (upper left index): 
\begin{eqnarray}
\prescript{L_{k-1}}{k-1}{H}^{}_{k} = ^0 L_{k-1}\inv {^0 L_{k}}
\label{eq:BFFIC}
\end{eqnarray}
Fig.~\ref{fig:Coordinates}a shows this transformations for three consecutive object poses. The new rigid body coordinates are given by the incremental pose transformation:
\begin{eqnarray}
{^0 L_{k}}  = {^0 L_{k-1}} \: \prescript{L_{k-1}}{k-1}{H}^{}_{k}
\label{eq:RelMotion}
\end{eqnarray}
Consider a point $^L l^i$ in the object frame $\{L\}$. Writing the expression \eqref{eq:RelPoint} for two consecutive poses of the object at time $k-1$ and $k$ and using the relative motion of the object in \eqref{eq:RelMotion}, the motion of this point can written as:
\begin{eqnarray}
^0\bl_{k}^i =\: ^0 L_{k-1}  \: \prescript{L_{k-1}}{k-1}{H}^{}_{k}  \:^0 L_{k-1}\inv \:  ^0\bl_{k-1}^i \:.
\label{eq:PointMotion}
\end{eqnarray}
We observe that \eqref{eq:PointMotion} relates the same point on the rigid body in motion at different time steps by a transformation \mbox{$\prescript{0}{k-1}{H}^{}_{k} = \:^0 L_{k-1} \:  \prescript{L_{k-1}}{k-1}{H}^{}_{k}\: ^0 L_{k-1}\inv $}, where \mbox{$\prescript{0}{k-1}{H}^{}_{k}\in \SE$}. According to ~\cite{Chirikjian17idetc}, this equation represents a frame change of a pose transformation, and shows how the body-fixed frame pose change in \eqref{eq:BFFIC} relates to the \emph{reference frame pose change}. The point motion in the reference frame becomes:
\begin{eqnarray}
\boxed{^0\bl_{k}^i =\: \prescript{0}{k-1}{H}^{}_{k} \: ^0\bl_{k-1}^i \:.}
\label{eq:ifTransform}
\end{eqnarray}
This formulation is key to the proposed approach since it eliminates the need to estimate the object pose $^0 L_{k}$ and allows us to work directly with points ${^0 \bl^i_k}$ in the reference frame.

\subsubsection*{Linear velocity extraction}
Other vehicle velocity is a crucial piece of information in autonomous driving applications. Given vehicle's rigid body pose change in inertial frame $\prescript{0}{k-1}{H}^{}_{k}$, its linear velocity vector in ($\frac{1}{\textrm{fps}}.\frac{\textrm{m}}{\textrm{s}}$) can be computed:
\begin{equation}
v = \prescript{0}{k-1}{t}^{}_{k} - (I_3- \prescript{0}{k-1}{R}^{}_{k})\: c_{k-1}
\label{eq:speed}
\end{equation}
and its speed is the magnitude of this vector, where $\prescript{0}{k-1}{R}^{}_{k} \in \SO$ and $\prescript{0}{k-1}{t}^{}_{k} \in \R^3$ the rotation and translation components of the vehicle's pose change in inertial frame $\prescript{0}{k-1}{H}^{}_{k}$ respectively, $I_3$ is the identity matrix, and $c_{k-1}$ is the object's centroid position at time $k-1$. As our algorithm is sparse, we do not have access to the object's centroid but rather approximate it by the 3D centroid of features detected on the object. The derivation of \eqref{eq:speed} is detailed in the appendix and for more explanation, we refer the reader to~\cite{Chirikjian17idetc}.

\begin{figure}[t]
	\begin{minipage}[t]{0.22\textwidth}
		\begin{minipage}[b]{\textwidth}
			\centering
			\includegraphics[height=2cm,width=1\linewidth,trim=50mm 205mm 85mm 40mm,clip]{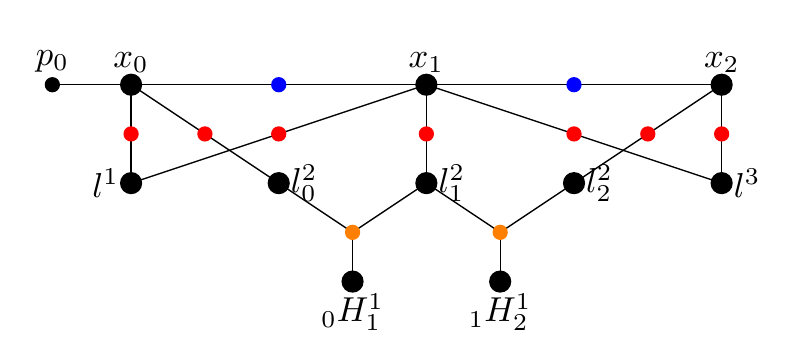}
			\\(a)
		\end{minipage}
	\end{minipage}\hspace{0.7cm}
	\begin{minipage}[t]{0.22\textwidth}
		\begin{minipage}[b]{\textwidth}
			\centering
			\includegraphics[height=2cm,width=1\linewidth,trim=50mm 205mm 85mm 40mm,clip]{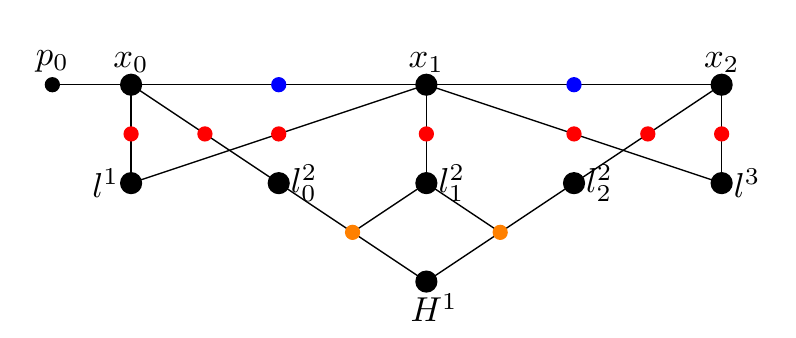}
			\\(b)
		\end{minipage}
	\end{minipage}
	\caption{\scriptsize Back-end (a): Factor graph representation of a problem with multiple pose change vertices for the same object. (b): Factor graph representation of a problem with a unique pose change vertex for the same object.}
	\label{fig:factorGraph}
\end{figure}

\subsection{Motion factors in dynamic SLAM}
The proposed approach estimates the camera poses, the static and dynamic structure and the motion of the dynamic structure. To achieve this, motion factors along with the odometry obtained from the robot's proprioceptive sensors, and the landmarks observations are optimised jointly:
\begin{flalign}\nonumber
\boldsymbol \theta^* =  \argmin_{\boldsymbol \theta} \Big\{ 
\sum_{k=1}^{m_k} \rho_{h}((h(x_{k},l^i_k) - z^i_k)\tr \Sigma_{w_k}\inv (h(x_{k},l^i_k) - z^i_k)) + \\\nonumber
\sum_{i=1}^{m_i} \rho_{h}((f(x_{k-1},x_k) - o_k)\tr \Sigma_{v_k}\inv (f(x_{k-1},x_k) - o_k)) + \\
\sum_{i,j}^{m_s} \rho_{h}((g(l^i_{k-1},l^i_{k},\prescript{0}{k-1}{H}_{k}^j)\tr  \Sigma_{q}\inv (g(l^i_{k-1},l^i_{k},\prescript{0}{k-1}{H}_{k}^j) 
\Big\} 
\end{flalign}
\label{eq:NLSMC}
where $\rho_{h}$ is the Huber function, \mbox{$h(x_k,l^i_k)$} is the 3D point measurement model with $\Sigma_{w_k}$ the point measurement covariance matrix; \mbox{$\textbf{z} = \{z_1 ... z_{m_k}\}$$|z_k \in {\rm I\!R}^3$} is the set of all ${m_k}$ 3D point measurements at all time steps, \mbox{$f(x_{k-1},x_{k})$} is the odometry model with $\Sigma_{v_k}$ odometry covariance matrix and \mbox{$\textbf{o} = \{o_1 ... o_{m_i}\} $} the set of ${m_i}$ odometric measurements. Fig.~\ref{fig:Coordinates}b shows the measurements in red. \mbox{$g(l^i_{k-1},l^i_{k},\prescript{0}{k-1}{H}_{k}^j)$} is the motion model of points on dynamic objects with $\Sigma_{q}$ motion covariance matrix and $m_s$ is the total number of motion factors. 
The motion of any point on a detected rigid object $j$ can be characterised by the same pose transformation $\prescript{0}{k-1}{H}_{k}^j \in \SE$ given by \eqref{eq:ifTransform} and the corresponding factor is:
%
\begin{eqnarray}
g(l^i_{k-1},l^i_{k},\prescript{0}{k-1}{H}^{j}_{k}) = { ^0l_{k-1}^i - \prescript{0}{k-1}{R}_{k}^j} \: ^0l_{k-1}^i - \: \prescript{0}{k-1}{t}_{k}^j  + q_{s_j}
\label{eq:MotionGeneral}
\end{eqnarray}
where  \mbox{$q_s \sim \mathcal{N}(0,\Sigma_{q})$} is the normally distributed zero-mean Gaussian noise. The factor in \eqref{eq:MotionGeneral} is a ternary factor which we call the \emph{motion model of a point on a rigid body} (orange factors in Fig.\ref{fig:factorGraph}). All the variables are grouped in ${\bm \theta} = {\bf x} \cup {\bf l} \cup{\bf H}$, where ${\bf H}$ is the set of all the variables characterising the objects' motions. 
\subsection{The factor graph}
\label{sec:factorGraph}

We model the dynamic SLAM problem as a factor graph. The factor graph formulation is highly intuitive and has the advantage that it allows for efficient implementations of batch~\cite{Dellaert06ijrr} \cite{ceres-solver} and incremental~\cite{Kaess11ijrr,Polok13rss,Ila17ijrr} solvers. 
It has been shown that in dynamic SLAM, knowing the type of motion of the objects in the environment is highly valuable~\cite{Wang07ijrr}. In this work we evaluate two scenarios without and with constant motion model :\\
%
%
$\bullet$ In city scenarios, where the objects motions are subject to changes (acceleration, deceleration, etc.) modelling the motion is challenging. Therefore, we allow for the estimation of a new pose change at every time step. Fig.~\ref{fig:factorGraph}a shows a factor graph representation of such scenario where the motion of the same object is estimated using two motions vertices for two different time transitions. A possible constraint is to minimise the change between these motion estimates.\\
$\bullet$ A highway scenario, where every vehicle maintains a constant motion. Fig.~\ref{fig:factorGraph}b shows the factor graph representation where a single motion is estimated per object. 

Further we show that if the body-fixed frame pose change is constant then the reference frame pose change is constant too. For any $k-1, {k'-1}$ time indices, the constant motion in the body-fixed pose change is:
\begin{eqnarray}
\prescript{L_{k-1}}{k-1}{H}^{}_{k} = C = \prescript{L_{k'-1}}{k'-1}{H}^{}_{k'} \in \SE\:.
\label{eq:ConstantC}
\end{eqnarray}
We rescale \eqref{eq:RelMotion} and use \eqref{eq:ConstantC} to obtain:
$^0 L_{k}  = ^0 L_{k-1}\, C$
which we replace in $\prescript{0}{k-1}{H}^{}_{k} = \:^0 L_{k-1} \:  C\: ^0 L_{k-1}\inv$ to obtain:
\begin{eqnarray}
\prescript{0}{k-1}{H}^{}_{k} = ^0 L_{k} \: C  \: ^0 L_{k}\inv =\: 
\prescript{0}{k}{H}^{}_{k+1}\:
\end{eqnarray}
It follows that the reference frame pose change for a specific object $j$:
$\prescript{0}{k-1}{H}_{k}^j = \prescript{0}{}{H}^j = 
\prescript{0}{k'}{H}_{k'+1}^j \in \SE$
holds for any $k, k'$ time indices and the factor in ~\eqref{eq:MotionGeneral} is changed accordingly.
The constant motion assumption can be used to handle occlusions by keeping hypothesis of previously detected objects and reviving those based on re-observations of occluded objects.
\section{System Overview}
\label{sec:System}
\begin{figure}[t]
	\centering
	\includegraphics[width=0.95\textwidth,trim=0cm 9cm 0cm 0.4cm,clip]{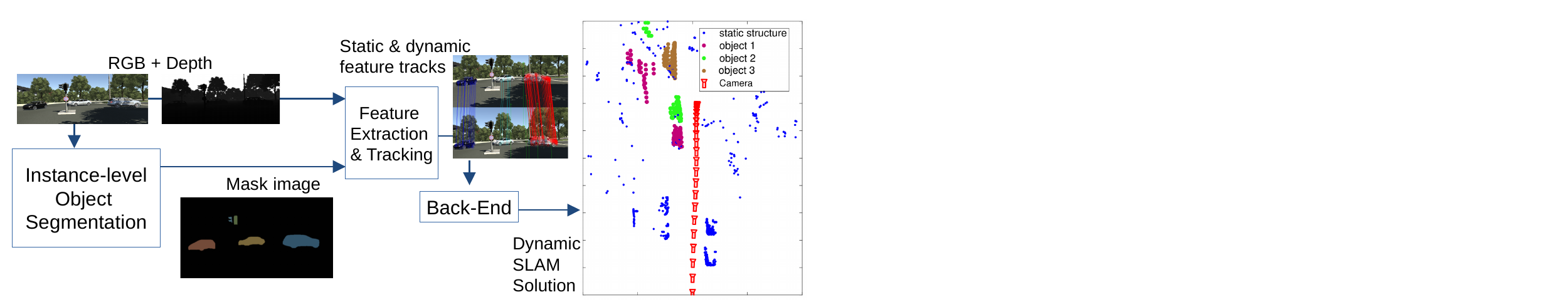}
	\caption{\scriptsize System overview. Input images are used into an instance level object segmentation algorithm to provide objects masks. The algorithm then detects and tracks features on potentially moving objects. Potentially dynamic features along with tracked static features are used to build the graph, that is then fed into a back-end optimisation.}
	\label{fig:system}
\end{figure}
The system pipeline shown in Fig.~\ref{fig:system} assumes a robot equipped with an RGB-D camera and proprioceptive sensors (e.g. odometers, IMU).
Our feature-based object-aware dynamic SLAM back-end estimates the robot poses, the static and dynamic structure and pose transformations for every detected object in the scene. To ensure features are being detected on moving objects, we employ an instance-level object segmentation algorithm to produce objects masks.
Object segmentation constitutes an important prior in static/dynamic object classification and tracking of dynamic objects. The front-end then makes use of object masks to detect features on \emph{potentially-moving} objects and on static background. Feature tracking is a crucial module for the success of our approach. Through object segmentation and feature tracking, the SLAM front-end is able to identify and associate points on the same rigid-body object at different time steps. These points share an underlying motion model that we exploit to achieve simultaneous localisation, mapping and moving object tracking. 
The algorithm does not require the front-end to estimate the objects' pose or use any geometric model of the objects. 
The static and dynamic 3D measurements along with the measurements from proprioceptive sensors are integrated into the back-end to simultaneously estimate the camera motion, the static and dynamic structure and the \SE pose transformations of detected objects in the scene.


\section{Experiments And Results}
\label{sec:Experiments}
\subsection{Error Metrics}
The accuracy of the solution is evaluated vs ground-truth (GT) by comparing the Relative Translational Error (RTE) in \%, that is the translational component of the error between the estimated and GT robot pose changes. Similarly, the Relative Rotational Error (RRE) in \textrm{$^\circ/m$} is the rotational component of the same error.
We also evaluate the Relative Structure Error (RSE) in \% for all static and dynamic landmarks, as the error between the corresponding relative positions of the estimated and GT structure points in the simulated experiments.
We also provide an evaluation of the object pose change estimates; the Object Motion Translation Error (OMTE) in \%, the Object Motion Rotational Error (OMRE) in \textrm{$^\circ/m$} and for driving scenarios, the Object Motion Speed Error (OMSE) in \%.

\subsection{Virtual KITTI Dataset}
\label{sec:vKITTI}
$\mathbf{Description:}$
Virtual KITTI~\cite{Gaidon16cvpr} is a photo-realistic synthetic dataset that provides \mbox{RGB-D} videos from a vehicle driving in an urban environment. Frames are fully annotated at the pixel level with unique object tracking identifiers (needed for errors calculations). GT information about camera and object poses is also provided which makes it a perfect dataset to test and evaluate the proposed technique. 

$\mathbf{Goal:}$
We make use of the GT data to test the effect of each component in the front-end on the performance of the algorithm and the accuracy of the pose change estimation for the camera and moving objects in the scene. The three aspects studied are errors in: \begin{inparaenum}[a)] \item depth/3D point measurement, \item object segmentation, and \item feature tracking\end{inparaenum}.

$\mathbf{Implementation:}$
Due to the fact that our algorithm is sparse-based, object pose change estimation is affected by the distribution of the extracted features on moving objects. 
Another important aspect is the percentage of the object mask in the image.
In the experiments reported in this paper, we only estimate for objects whose segmentation masks amount to a certain percentage of the total image. This threshold ensures to exclude far-away and partially observed objects that are entering/exiting the camera field of view and which makes their motion estimates inaccurate. This threshold is set to 6\% for vKITTI, and 2\% for KITTI.

\begin{figure}[t]
	\centering
	\includegraphics[width=0.45\textwidth,trim=0.2cm 0.2cm 0cm 0cm,clip]{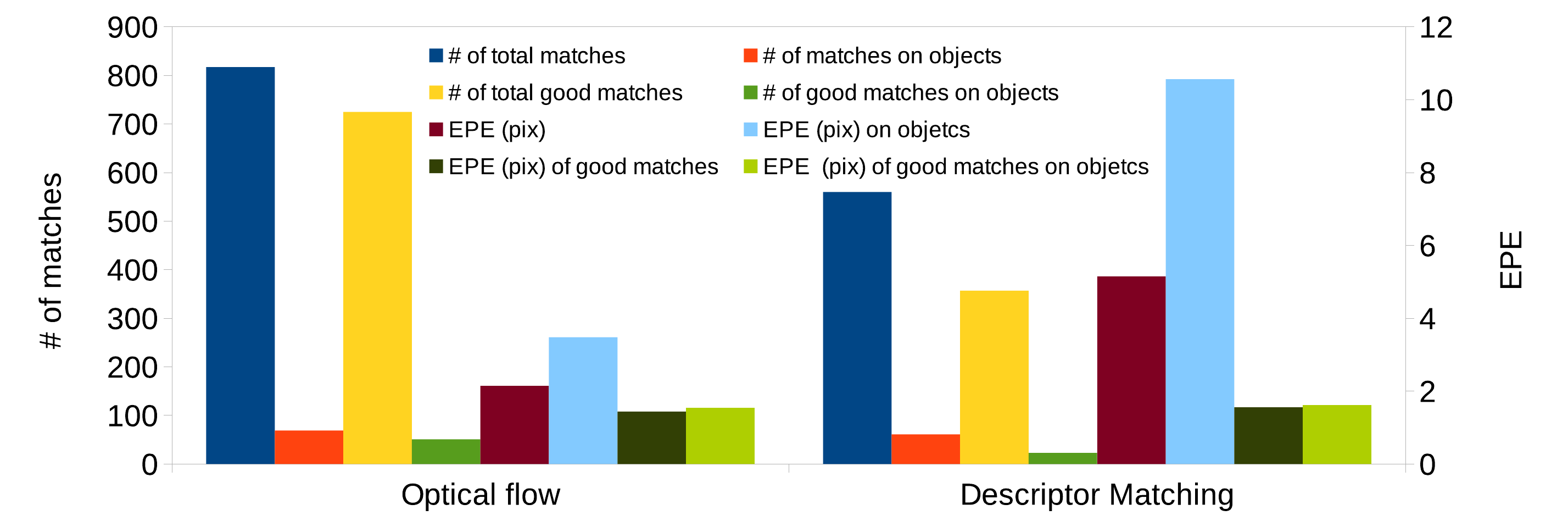}
	\caption{{\scriptsize Comparison of optical flow and descriptor matching for feature tracking.}}
	\label{fig:flowVsMatching}
\end{figure}

\paragraph{Depth error}
\label{sec:baseline}
We evaluate the performance of our algorithm using GT object segmentation, and feature tracking with odometry and varied point measurement noise. The noise levels added are 5\% for translational odometry in each axis, and 10\% for rotational odometry around each axis. Three different noise levels, drawn from a normal distribution with zero mean and standard deviation  $\sigma_1$=0.02, $\sigma_2$=0.04, and $\sigma_3$=0.06\ m in each axis per observation, were tested. These noise levels correspond to commercially available LiDAR system, and a stereo-camera rig respectively and a third higher value.
This is conceptually the same as replacing the depth input in the front-end with a stereo depth estimation algorithm e.g. SPSS~\cite{Yamaguchi14cvpr} or a single image depth estimation for a monocular system e.g.~\cite{Ren19arXiv}. Point measurement noise is kept at $\sigma_1$ for further tests in this subsection.
\paragraph{Object segmentation error}
This test aims at evaluating the effect of the object segmentation while using GT feature tracking with added odometry and point measurement noise. We employ MASK-RCNN~\cite{He17iccv}, learning based model, for instance-level object segmentation. We perform evaluation tests of MASK-RCNN on all sequences of vKITTI and KITTI. Results for mean average precision (mAP) and mean intersection over union for predictions only (mIOU\_Pred) of the `car' class are 0.513 and 0.557 for vKITTI and 0.413 and 0.632 for KITTI. Numbers show good performance, however, testing the effect on camera and object pose change estimation is crucial. 
\paragraph{Feature tracking error}
As our algorithm is sparse feature-based, feature tracking is an essential component of the front-end. In order to test the effect of feature tracking, we first conduct tests on the quantity and quality of feature matches using 1) PWC-Net~\cite{Sun18cvpr} and 2) feature descriptor matching. Fig.~\ref{fig:flowVsMatching} shows the number of total and object matches and their corresponding end-point error (EPE), and then extends this test to show these values for ``good matches"; matches with less than 3 pixels EPE.

\begin{figure}[t]
 \centering
 \includegraphics[width=0.48\textwidth,trim=0.5cm 0.5cm 0.5cm 0cm,clip]{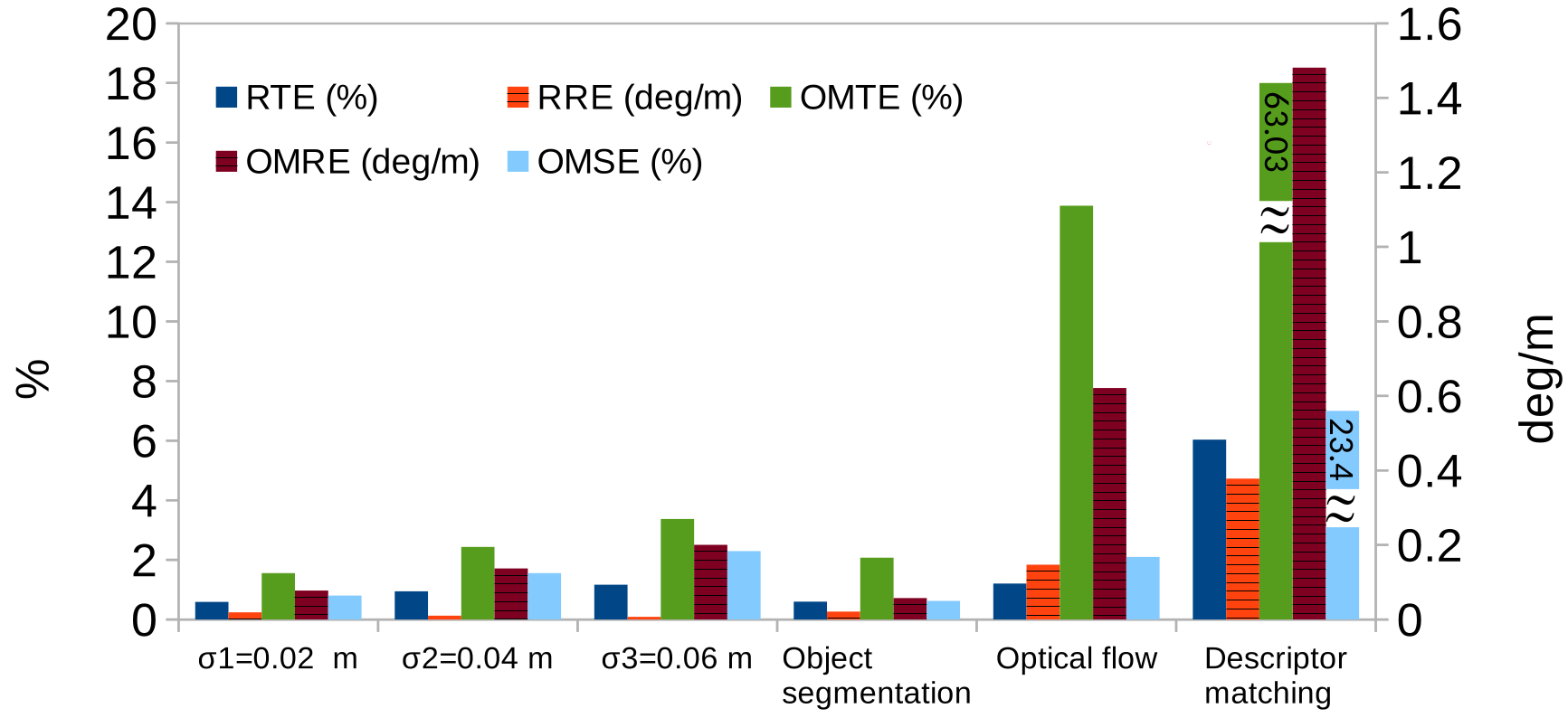}
 \caption{{\scriptsize Study evaluated on vKITTI of the effect of different front-end components on the camera/objects pose change estimation accuracy.}}
 \label{fig:ablationStudy}
\end{figure}

$\mathbf{Discussions:}$
As shown in Fig.~\ref{fig:ablationStudy}, the feature tracking is the most crucial component of the front-end and dictates the performance of our feature-based algorithm.
Fig.~\ref{fig:flowVsMatching} and ~\ref{fig:ablationStudy} show better performance of optical flow over descriptor matching in terms of quantity and quality of features.
For the remainder of this paper, optical flow is used for feature tracking and we aim to look at improving the feature tracking in a future work by utilising the object motion estimates.
Object segmentation appears to have the least effect on the estimation quality. Errors in the camera and object motion estimation due to the use of MASK-RCNN compared to GT segmentation appear to be minimal.
\subsection{Simulated Data}
\label{sec:simulatedData}

$\mathbf{Description:}$
This experiment features a single simulated ellipsoid-shaped object tracked by a robot as it follows a circular motion in an environment with no static structure. The object is simulated to have a constant \SE pose change, and the estimation makes use of this piece of information to constraint the problem as explained in Subsection~\ref{sec:factorGraph}. The simulation corresponds to a scenario where only moving structure is visible, e.g. a vehicle on a bridge or inside a tunnel occluded by other vehicles driving alongside and failing to track static structure.

$\mathbf{Goal:}$
This experiment is designed to show that our approach provides good solutions in cases where existing approaches to dynamic SLAM might fail. We compare our algorithm (one framework joint estimation) vs. parallel tracking and mapping, e.g. SLAM + Multiple Object Tracking (MOT)~\cite{Wang03icra,Miller07icra,Rogers10iros,Kundu11iccv}. This class of algorithms depend on the quality of the returned map, and will perform poorly in environments with insufficient number of reliable static structure such as the examples given above.  
%
%
%
\begin{table}[t]
	\normalsize
	\centering
	\caption{\small Results of applying our object-aware dynamic motion integration on a simulated data}
	\label{tab:resultsSimulated}
	\resizebox{0.7\columnwidth}{!}{%
	\begin{tabular}{c C{2.5cm} C{2.5cm}}
		\Xhline{1pt}
		\textbf{Error} & \textbf{\makecell{SLAM+MOT}} & \textbf{\makecell{Ours}}
		\\ \hline
		RTE (\%) & 4.426 & \textbf{3.804} \\ 
		RRE (\textrm{$^\circ/m$}) & 1.34 & \textbf{0.486} \\ 
		RSE (\%) & 8.019 & \textbf{4.177} \\ 
		OMTE (\%) & 20.946 & \textbf{4.018}\\ 
		OMRE (\textrm{$^\circ/m$}) & 0.349 & \textbf{0.055} \\ \Xhline{1pt}
	\end{tabular}
	}
\end{table}

$\mathbf{Discussions:}$
Camera motion in the case of parallel SLAM+MOT is basically a direct integration of odometric measurements. Results in Table~\ref{tab:resultsSimulated} show the clear advantage of our algorithm that jointly estimates the camera and rigid object pose transformations. Improvements are in the range of 80-85\% in object pose change estimation. In an extreme case, where no static structure is observed, our algorithm not only improves the object motion estimates but also the camera pose estimation. However, in an environment with enough static structure, both algorithms yield very similar results as shown in the next section.


\subsection{KITTI dataset}
\label{sec:KITTI}

\begin{table}[t]
	\normalsize
	\centering
	\caption{\small Results of applying our object-aware dynamic motion integration on KITTI}
	\label{tab:resultsKITTI}
	\resizebox{\columnwidth}{!}{%
	\begin{tabular}{c C{1.2cm}| C{1.2cm} C{1.2cm}| C{1.2cm} C{1.2cm}| C{1.2cm} C{1.2cm}| C{1.2cm} C{1.2cm}| C{1.2cm}}
		& \multicolumn{1}{c} \textbf{\makecell{Seq.007}} & \multicolumn{2}{c} \textbf{\makecell{Seq.006}} & \multicolumn{2}{c} \textbf{\makecell{Seq.0001}} & \multicolumn{2}{c} \textbf{\makecell{Seq.0003}} & \multicolumn{2}{c} \textbf{\makecell{Seq.0005}} & \makecell{Seq.0000} \\ \Xhline{1pt}
		\textbf{Error} & \textbf{\makecell{Static \\ Only}} & \textbf{\makecell{Static \\  Only}} & \textbf{\makecell{Ours}} & \textbf{\makecell{Static \\ Only}} & \textbf{\makecell{Ours}} & \textbf{\makecell{Static \\Only}} & \textbf{\makecell{Ours}} & \textbf{\makecell{SLAM + \\ MOT}} & \textbf{\makecell{Ours}} & \textbf{\makecell{Ours}} \\ \Xhline{1pt}
		RTE (\%) & 0.039 & 0.000 & 0.000 & 0.248 & 0.248 & 0.016 & 0.016 & 0.022 &  0.025 & 0.020\\ 
		RRE (\textrm{$^\circ/m$}) & 0.003 & 0.001 & 0.001 & 0.017 & 0.017 & 0.002 & 0.002 & 0.003 & 0.003 & 0.001\\ 
		OMTE (\%) & -- & -- & 11.646 & -- & 10.525 & -- & 6.0/59.5 & 23.608 & 23.653 & 42.7/63\\ 
		OMRE (\textrm{$^\circ/m$}) & -- & -- & 0.254 & -- & 0.555 & -- & 0.2/1.0 & 0.472 & 0.473 & 1.2/5.0\\ 
		OMSE (\%) & -- & -- & 10.587 & -- & 5.561 & -- & 2.5/2.7 & 11.809 & 11.809 & 20.7/22 \\\Xhline{1pt}
	\end{tabular}
}
\end{table}

\begin{figure*}[t]
	\begin{minipage}[t]{0.3\textwidth}
		\begin{minipage}[b]{\textwidth}
			\centering
			\includegraphics[width=1\linewidth]{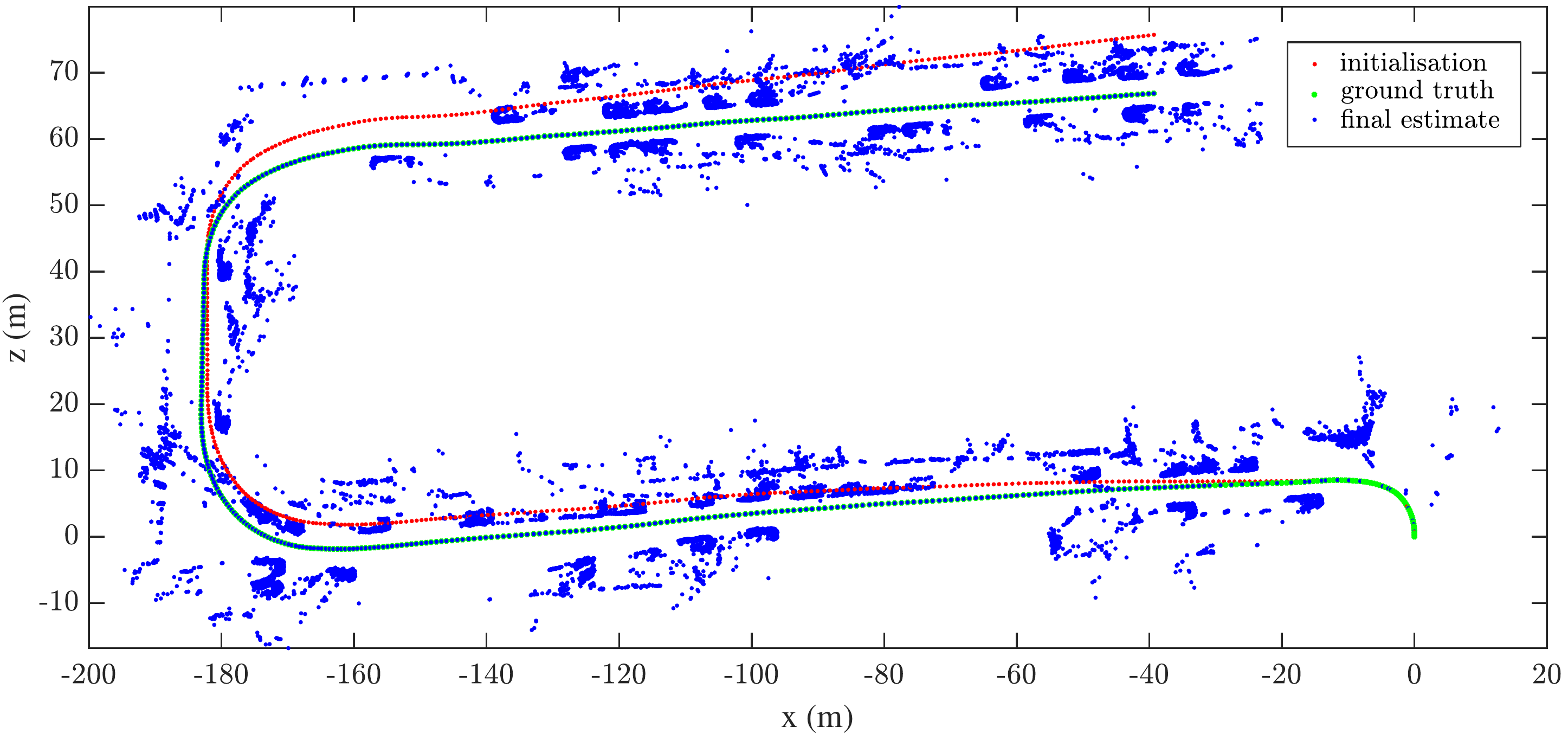}
			\\ (a) \scriptsize Sequence0007 frames 15-510. Static SLAM.
		\end{minipage}
		\label{fig:kitti7}
	\end{minipage}\hspace{0.3cm}
	\begin{minipage}[t]{0.3\textwidth}
		\begin{minipage}[b]{\textwidth}
			\centering
			\includegraphics[width=1\linewidth]{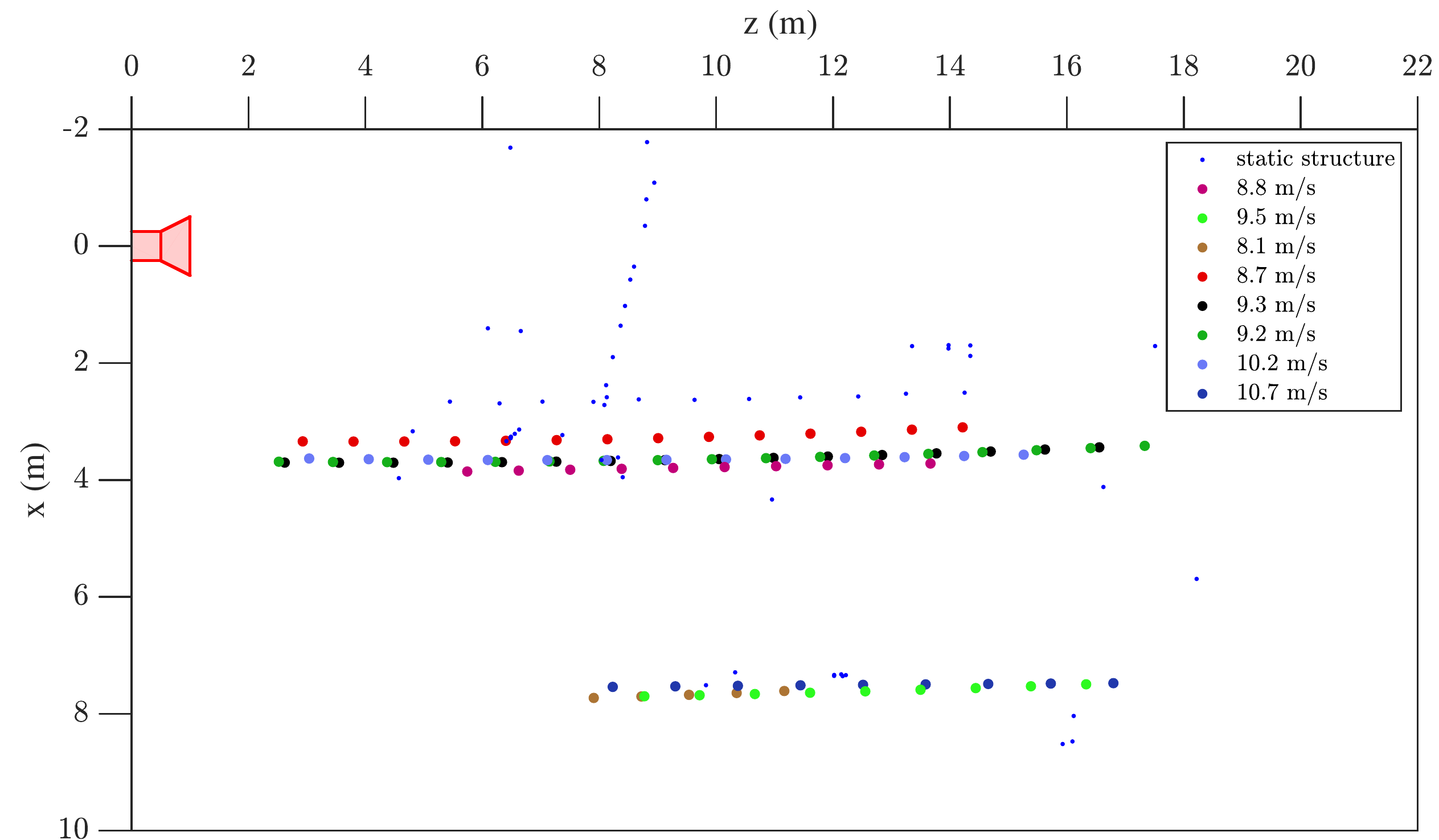}
			\\(b) \scriptsize Sequence0006 frames 40-140. MOT with a static camera.
		\end{minipage}
		\label{fig:kitti6}
	\end{minipage}
	\begin{minipage}[t]{0.3\textwidth}
		\begin{minipage}[b]{\textwidth}
			\centering
			\includegraphics[width=0.7\linewidth,trim=3cm 7.6cm 2cm 8cm,clip]{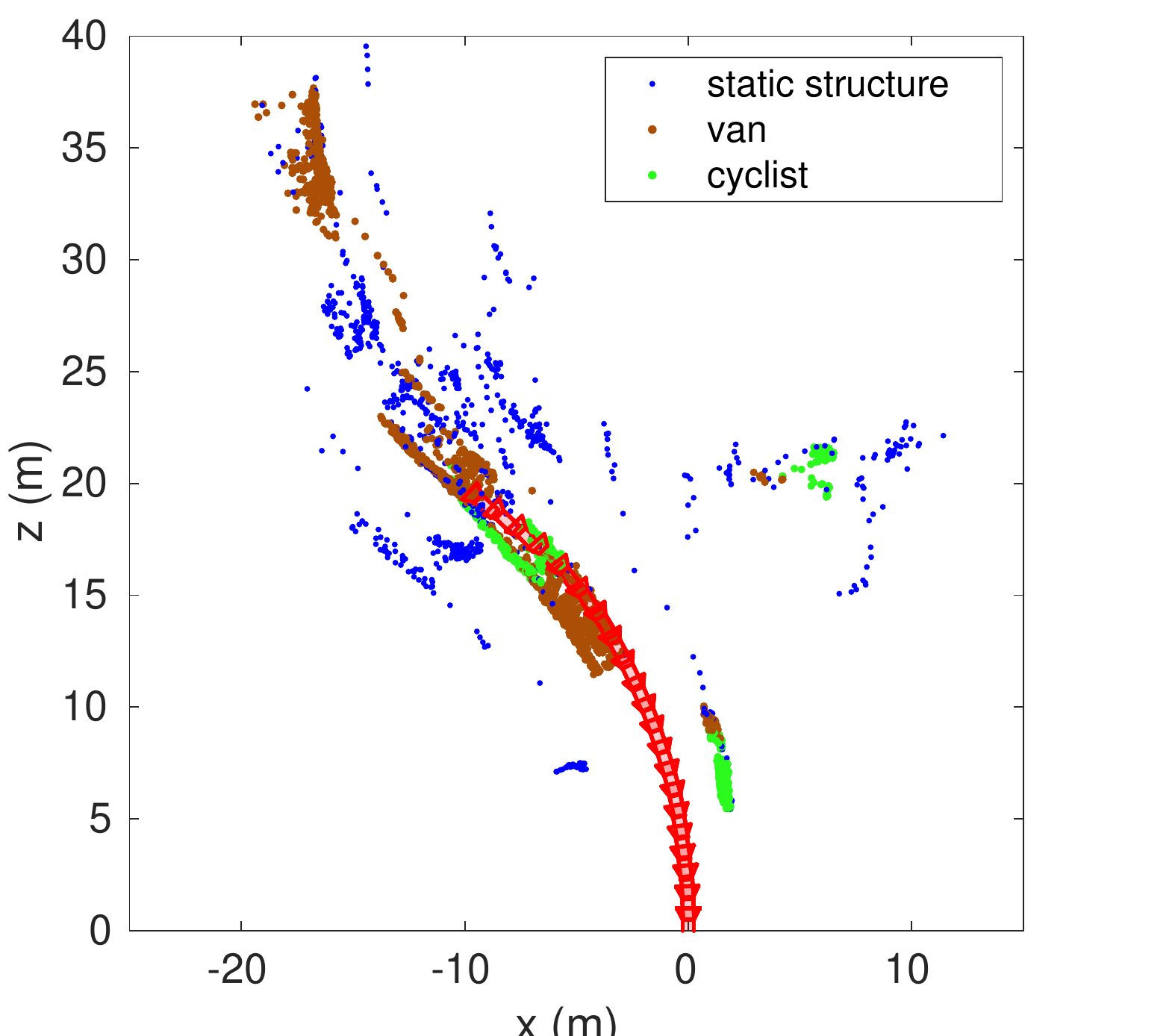}
			\\(c) \scriptsize Sequence0000 frames 0-60. Dynamic SLAM.		
		\end{minipage}
		\label{fig:kitti0}
	\end{minipage}
	\begin{minipage}[t]{0.45\textwidth}
		\begin{minipage}[b]{\textwidth}	
			\centering
			\includegraphics[width=1\linewidth,trim=0cm 0cm 0.3cm 0cm,clip]{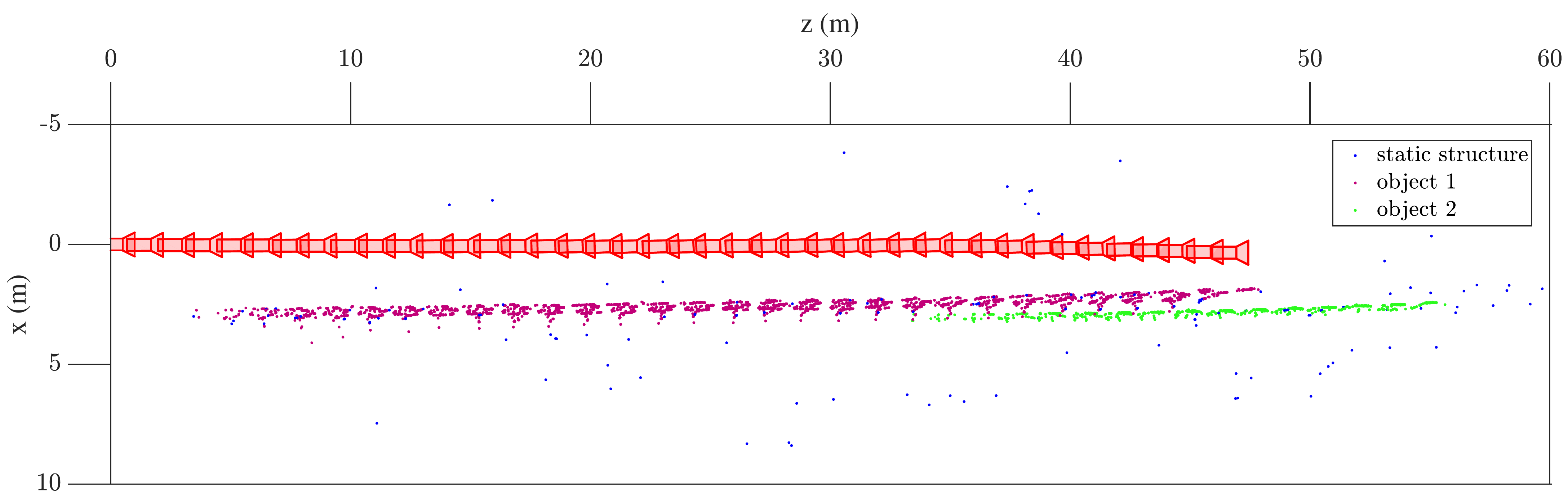}
			\\(d) \scriptsize Sequence0003 frames 0-40 results. Dynamic SLAM.
		\end{minipage}
		\label{fig:kitti3}
	\end{minipage}\hfill
	\begin{minipage}[t]{0.45\textwidth}
		\begin{minipage}[b]{\textwidth}
				\centering
				\includegraphics[height=2.6cm,width=1\linewidth,trim= 0.3cm 1cm 0cm 0cm,clip]{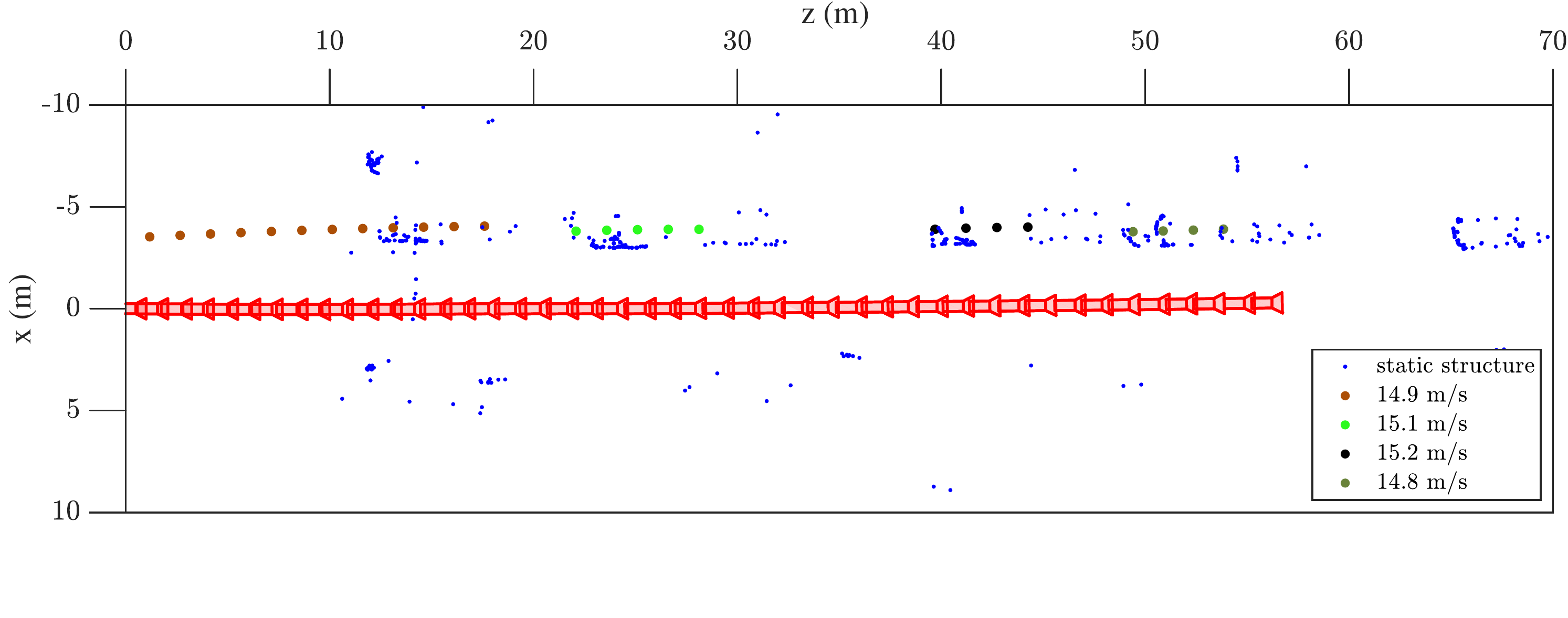}
				(e) \scriptsize Sequence0005 frames 185-229 results. Dynamic SLAM.		
		\end{minipage}
	\label{fig:kitti5}
	\end{minipage}
	\caption{\scriptsize Sample results on KITTI various sequences.}
	\label{fig:kitti}
\end{figure*}


$\mathbf{Description:}$
KITTI~\cite{Geiger12cvpr} has been a standard benchmark suite for a number of challenging real-world computer vision tasks. We make use of the KITTI tracking dataset as it provides GT object poses in camera coordinate frame.


$\mathbf{Goal:}$
This experiment is designed to evaluate the performance of our algorithm on real-world challenging outdoor scenarios. In relevant dataset sequences, we compare our results with classical static only SLAM (where dynamic objects are considered outliers) and SLAM+MOT solutions.

To demonstrate the generality of the approach and the fact that the proposed framework performs well in any type of scenarios, we consider three different cases:
\\
a) \emph{Classical SLAM}: A moving robot equipped with an RGB-D camera in a static environment. Sequence0007 represents this case and shows that our algorithm performs equally well in a classical scenario with no dynamics and requires no prior knowledge or makes any prior assumptions of the scene.
\\
b) \emph{Multi-object tracking}: A static camera in a dynamic environment as shown in Sequence0006. For this specific data sequence, we consider a constant pose change assumption. Note that camera pose change errors for this sequence are reported in meters and degrees.
\\	
c) \emph{Dynamic SLAM}: A moving robot equipped with an RGB-D camera in a dynamic environment. In here we do the distinction between two sub-scenarios:
\\ $\bullet$ A highway scenario, represented by Sequence0005, where every vehicle is assumed to have constant motion. This allows us to constraint the problem by assuming a constant pose change model for each detected object in the scene.
\\  $\bullet$ Sequence0001, Sequence0003 and Sequence0000 represent an intersection and other city driving scenarios, where motion models are difficult to impose. In here, the factor graph formulation allows for the estimation of a new pose change vertex every time step. Some insights on how to improve the estimation in such scenarios is provided in Section~\ref{sec:Conclusion}. Sequence0003 and Sequence000 contain two objects each, therefore the object motion error results are shown separated by a `/'.
Sequence0000 consists of a ``van'' and a ``cyclist'' which slightly violates the rigidity assumption, yet our approach still provides fairly good results.

$\mathbf{Implementation:}$
The three variants of SLAM: classical, SLAM+MOT, and dynamic are run using the front-end introduced in Section ~\ref{sec:System} and implemented in GTSAM~\cite{Dellaert12gtsam}. The tracking dataset is thought for camera-only based application, therefore GPS and IMU measurements are fused and further corrupted with noise to simulate odometric measurements available in a robotic (self-driving cars) scenario. The noise values are the same as the ones explained in Subsection~\ref{sec:baseline}, except Sequence0007, where twice the noise is added.
%
%
In autonomous driving, the literature normally distinguishes between different depth ranges for velocity estimation~\cite{Kampelmuhler18cvww}: near (d $<$ 20 m), medium (20 m $\leq$ d $<$ 45 m) and far range (d $>$ 45 m). In all KITTI experiments presented here, we only consider objects $<$ 22 m of distance to the camera (near and early medium range).

$\mathbf{Discussions:}$
All results show high accuracy in the estimation of pose change transformations and speeds of moving objects. Results in Table~\ref{tab:resultsKITTI} show a speed estimation accuracy in the range of 78-97.5\%. The second objects in Sequence0003 and Sequence0000 are particularly hard to process. In Sequence0003, the second object only occupies a small part of the image, dominated by its wheels having a different motion than the vehicle, yet its speed estimate is reasonable. Sequence0000 consists of a van and a cyclist turning at very low speeds ($<$ 5.5 m/s). Their motion estimation is particularly hard because of association errors and the fact that a cyclist is a non-rigid object mostly formed by wheels not obeying the motion model of the object.
%
%
Although speed errors seem high in percentage, they only account for an average speed error of 0.16 m/s for the van and 0.063 m/s for the cyclist.
\section{Conclusion And Future Work}
\label{sec:Conclusion}
In this paper we proposed a novel framework that exploits semantic information in the scene with no additional knowledge of the object pose or geometry, to achieve simultaneous localisation, mapping and tracking of dynamic objects. The algorithm shows consistent, robust and accurate results in various scenarios. Although the method presented here is applied to RGB-D/stereo images, we plan to explore semantic depth from single image or a purely monocular setup in the future. An important issue to be analysed, is the computational complexity of SLAM with dynamic objects. In long-term applications, different techniques can be applied to limit the growth of the graph~\cite{Strasdat11iccv,Ila10tro}. The estimation could be further enhanced by assuming a constant motion within a temporal window and use this assumption to handle occlusions and reduce the problem size.
Another possible extension is to use the SLAM back-end estimates to improve the tracking accuracy of the front-end.
\section*{Appendix}
\label{sec:appendix}
To see that~\eqref{eq:speed} is the same quantity in 3D as the translation vector from the origin of the object pose at time $\{k-1\}$ to the origin of the object pose at time $\{k\}$ as seen in $\{0\}$, we start by writing the object pose change in $\{0\}$ and substitute for $^{L_{k-1}}_{k-1}H_{k}$ by its definition in~\eqref{eq:BFFIC}
\begin{equation}
\prescript{0}{k-1}{H}^{}_{k} = \:^0 L_{k-1} \:  \prescript{L_{k-1}}{k-1}{H}^{}_{k}\: ^0 L_{k-1}\inv = \:^0 L_{k} \: ^0 L_{k-1} \inv
\end{equation}
Assuming $^0 R_{L_{k-1}} \in \SO$ and $^0t_{L_{k-1}} \in \R^3$ the rotation and translation components of $^0L_{k-1}$, and $^0 R_{L_{k}}$, $^0t_{L_{k}}$ their corresponding at time $k$, the translation and rotation parts of $\prescript{0}{k-1}{H}^{}_{k}$ can be expressed as $^0t_{L_{k}} - ^0R_{L_{k}} \: ^0 R_{L_{k-1}}\tr \: ^0t_{L_{k-1}}$ and $^0R_{L_{k}} \: ^0 R_{L_{k-1}}\tr$.
Substituting these two quantities into~\eqref{eq:speed}, we get
\begin{equation}
v =  \: ^0t_{L_{k}} - \: ^0R_{L_{k}} \: ^0 R_{L_{k-1}}\tr \: ^0t_{L_{k-1}} - (I_3 - \: ^0R_{L_{k}} \: ^0 R_{L_{k-1}}\tr) \: ^0t_{L_{k-1}}
\end{equation} which reduces to \mbox{$v = \: ^0t_{L_{k}} - \: ^0t_{L_{k-1}}$} which is the translation vector from the origin of the object pose at time $\{k-1\}$ to the origin of the object pose at time $\{k\}$ as seen in $\{0\}$.
\section*{ACKNOWLEDGMENT}
This research was supported by the Australian Research Council through the ``Australian Centre of Excellence for Robotic Vision'' CE140100016.
%
%
\bibliographystyle{IEEEtran}
\bibliography{biblio}


\end{document}